\documentclass[11pt]{article} 
\usepackage{wrapfig}
\usepackage{rldmsubmit}
\usepackage{palatino}
\usepackage{graphicx}
\usepackage{amsmath}
\usepackage{amsthm}
\usepackage{amssymb}
\usepackage{subfigure}
\usepackage{biblatex}
\usepackage[font=small]{caption, subcaption}
\title{Mapping Neural Signals to Agent Performance, A Step Towards Reinforcement Learning from Neural Feedback}

\author{
Julia Santaniello \\
Department of Computer Science\\
Tufts University\\
Medford, MA 02155 \\
\texttt{julia.santaniello@tufts.edu} \\
\And
Matthew Russell\\
Department of Computer Science\\
Tufts University\\
Medford, MA 02155 \\
\texttt{mrussell@cs.tufts.edu} \\
\And
Benson Jiang\\
Department of Computer Science\\
Tufts University\\
Medford, MA 02155 \\
\texttt{benson.jiang@tufts.edu} \\
\And
Donatello Sassaroli\\
Department of Computer Science\\
Tufts University\\
Medford, MA 02155 \\
\texttt{donatello.sassaroli@tufts.edu} \\
\And
Robert Jacob\\
Department of Computer Science\\
Tufts University\\
Medford, MA 02155 \\
\texttt{jacob@cs.tufts.edu} \\
\And
Jivko Sinapov\\
Department of Computer Science\\
Tufts University\\
Medford, MA 02155 \\
\texttt{jivko.sinapov@tufts.edu} \\
}

\begin{document}

\maketitle

\begin{abstract}
Implicit Human-in-the-Loop Reinforcement Learning (HITL-RL) is a methodology that integrates passive human feedback into autonomous agent training while minimizing human workload. However, existing methods often rely on active instruction, requiring participants to teach an agent through unnatural expression or gesture. We introduce \emph{NEURO-LOOP}, an implicit feedback framework that utilizes the intrinsic human reward system to drive human-agent interaction. This work demonstrates the feasibility of a critical first step in the \emph{NEURO-LOOP} framework: mapping brain signals to agent performance. Using functional near-infrared spectroscopy (fNIRS), we design a dataset to enable future research using passive Brain-Computer Interfaces for HITL-RL. Participants are instructed to observe or guide a reinforcement learning agent in its environment while signals from the prefrontal cortex are collected. We conclude that a relationship between fNIRS data and agent performance exists using classical machine learning techniques. Finally, we highlight the potential that neural interfaces may offer to future applications of human-agent interaction, assistive AI, and adaptive autonomous systems.
\end{abstract}


\acknowledgements{We would like to acknowledge the members of the MuLIP and HCI labs for their support, specifically Kenny Zheng, Anes Kim, Anna Sheaffer, Brennan Miller-Klugman, and Iris Yang.}

\startmain 

\section{Introduction}
\vspace{-5pt}
Recent advancements in human-in-the-loop reinforcement learning (HITL-RL) have become crucial to training and fine-tuning state-of-the-art systems \cite{wu2023finegrained, knox_augmenting_2011}. However, these feedback methods often require active, focused participation or expert demonstration for more complex tasks \cite{liRLsurveyRLHF}.

\begin{wrapfigure}{r}{0.4\textwidth}
\centering
\includegraphics[width=1.0\linewidth]{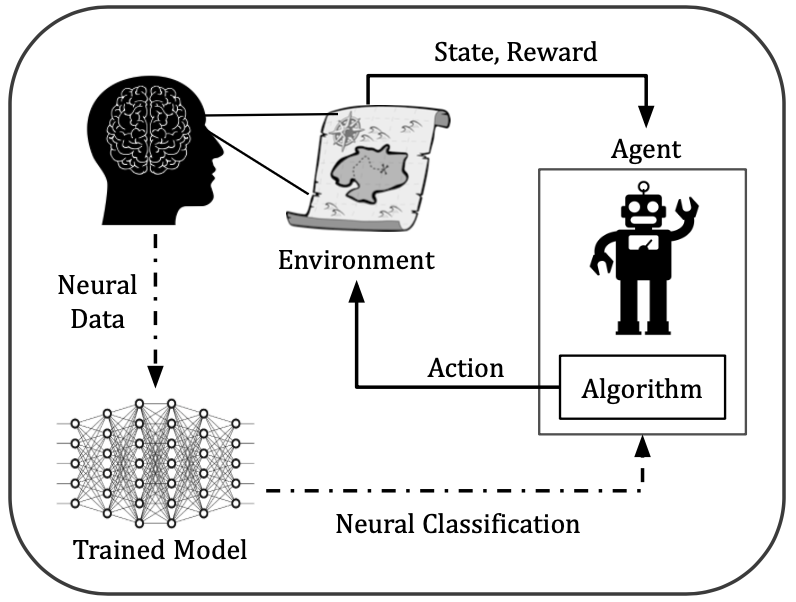}
\caption{\textmd{\emph{NEURO-LOOP} Diagram}}
\label{fig:framework-diagram}
\end{wrapfigure}
 
 Implicit human feedback systems present a valuable avenue for circumventing these limitations. Current methods leverage facial expression or gesture classification, allowing enhanced agent training while reducing cost to the human teacher \cite{Arakawa2018DQNTAMERHR, cui_empathic_2021}. Occasionally, these experimental procedures are partially explicit, requesting participants to actively teach the agent through their reactions. Finally, these models rely on data that can be difficult to access, collect or distribute, creating technical barriers for research.

Passive Brain-Computer Interfaces (BCI) offer a unique interface for HITL-RL frameworks while circumventing some of these limitations. Human neural feedback can be both fully implicit and easily distributed. Functional near-infrared spectroscopy (fNIRS) provides a useful brain signal, a measurement of hemodynamic response in the prefrontal cortex (PFC). Previous work has found existing relationships between hemodynamic response and cognitive or mental workload \cite{huangfNIRS2MW2021, wang_taming_2021}. The intrinsic “reward function” inherent in the human dopaminergic system has led to past work that suggests a correlation between major game-play events and fNIRS neural data \cite{fNIRSRewardDanceSimulation2012, VideoGamefNIRS2018, MorelliVicariousPersonal2015}. Other studies have researched a similar framework by utilizing binary error-related potentials (ErrP) to enhance agent training \cite{xu_accelerating_2021, agarwal_human---loop_2020}.

We hypothesize that an adjacent relationship exists between the human hemodynamic response and agent performance. Our results suggest that human brain signals can be mapped to agent performance, and richer signal granularity can be extracted, highlighting a critical first step in creating an implicit, brain-driven HITL-RL framework.

\section{Dataset}
\vspace{-5pt}
We formalize the structure of the complete dataset as $\mathcal{D}$, and the subset of data from each participant $k$, may be denoted as $\mathcal{D}^k$. Dataset $\mathcal{D}^k$ further divides into two subsets, the neural data set and the task data set.

\textbf{Neural Data Set} $\vert$ We define the neural dataset $\mathcal{N}^k$ as the set of neural channel recordings over time for some participant $k$. Let $\mathcal{N}_i \in \mathbb{R}^{M \times T}$ denote the neural signal data matrix at instance $i$, where $M$ is the number of neural channels and $T$ is the number of timestamps. A single neural data vector at timestamp $t$ across all channels can be denoted as $\mathbf{x}_t \in \mathbb{R}^M$. The signal at a specific timestamp $t$ and channel $m$ can be expressed as $x_{tm}$, where $t \in \{1, \dots, T\}$ and $m \in \{1, \dots, M\}$.

\textbf{Task Data Set} $\vert$ We define the Task Dataset $\mathcal{H}^k$ as the set of learning task statistics over time for some participant $k$. Let $\mathcal{H}_i \in \mathbb{R}^{P \times T}$ represent the task data matrix at instance $i$, where $P$ is the number of task variables and $T$ is the number of timestamps.

We may describe a vector of task variables/statistics as 
$
\mathcal{H}_k = \{S_t, A_t, R_t, S_{t+1}, V_{t}, B_{t}, E_{t}\}_{t=0}^{t=T},
$
where $S_t$ is the agent's state at time $t$, $A_t$ is the action chosen by the agent or human, $R_t$ is the reward, $S_{t+1}$ is the next state, and $V_{t}, B_{t}, E_{t}$ represent agent optimality values. An instance of a task data point at some timestamp for some specific statistic type can be denoted as $h_{tp}$, where $t \in \{1, \dots, T\}$ and $p \in \{1, \dots,P\}$.

$V \in \mathcal{R}^T$ is used to denote a set of discrete classification labels, where $v_t$ is a discrete number that correlates to agent performance or optimality at time $t$. Similarly, we use $B$ to denote a set of binary labels that correlates to agent performance. 
\begin{wrapfigure}{R}{.2\textwidth}
    \begin{minipage}{\linewidth}
    \centering{justification=centering}
    \includegraphics[width=0.8\linewidth]{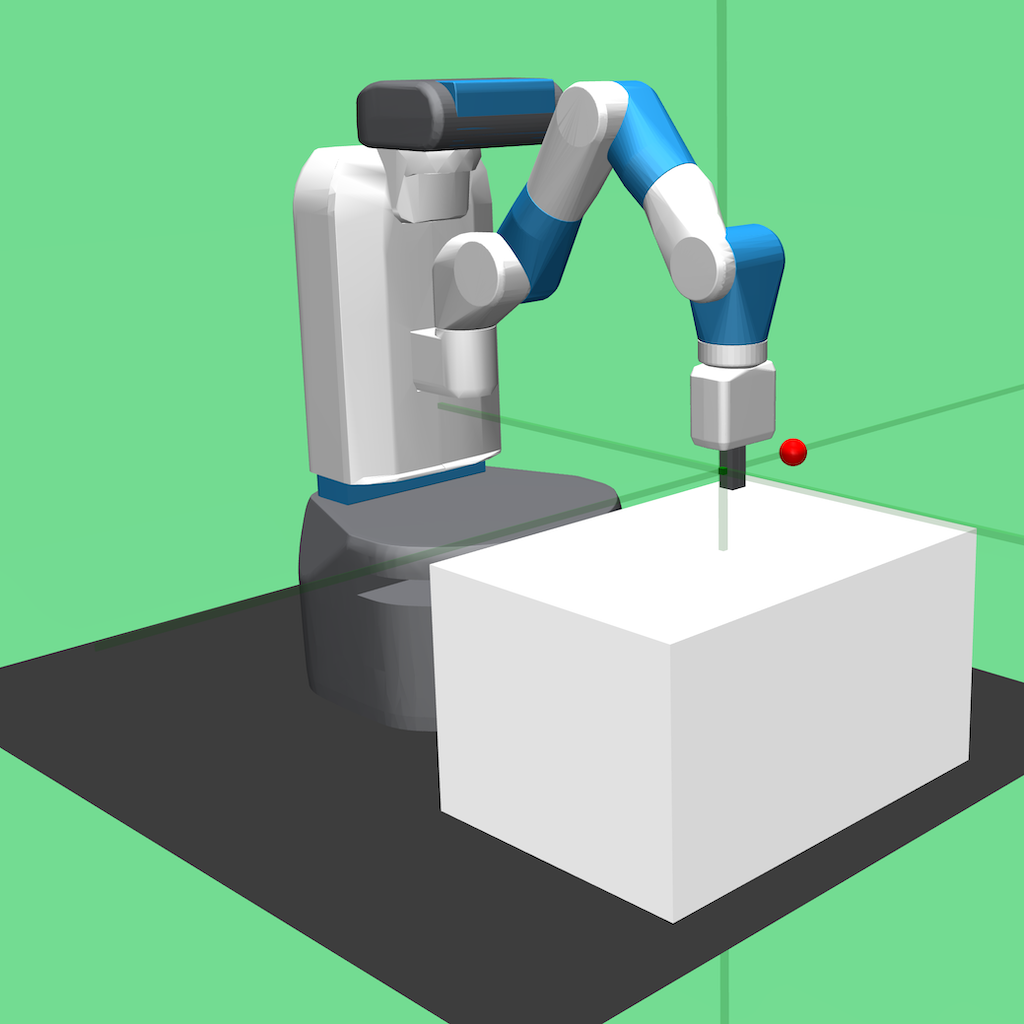}
    \includegraphics[width=0.8\linewidth]{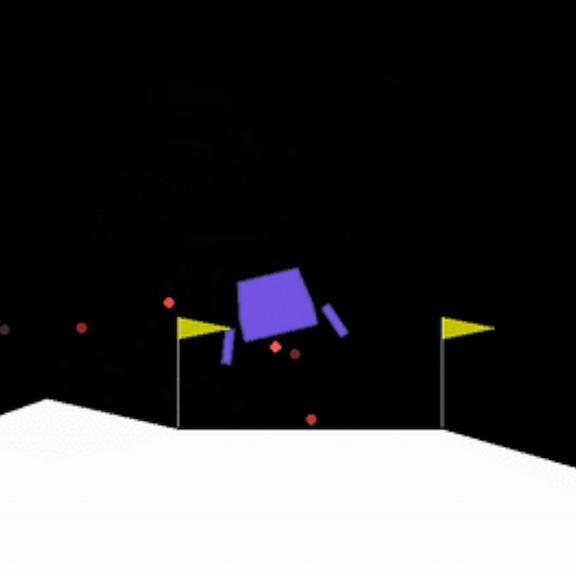}
    \includegraphics[width=0.8\linewidth]{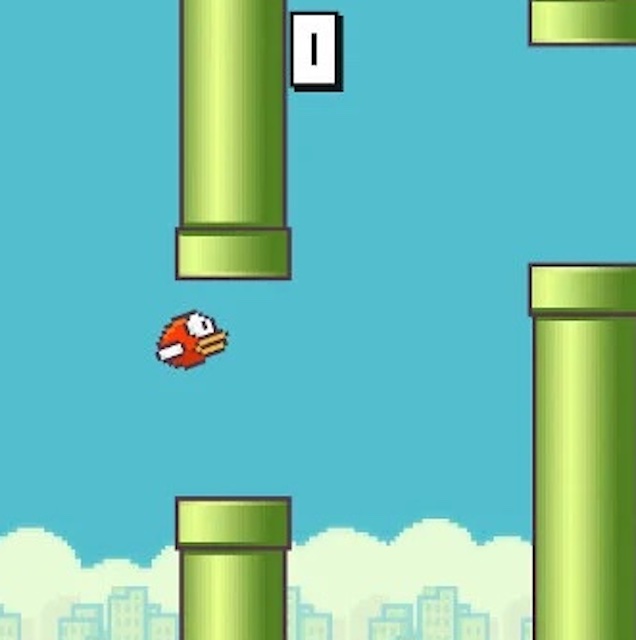}
\end{minipage}
\caption{OpenAI Gymnasium environments.}\label{fig:5}
\vspace{-\baselineskip}
\end{wrapfigure} 
Finally, $E$ is used denote a set of continuous optimality labels, where $e_t$ is a continuous number that represents the error between an agent's chosen action and the chosen action of a set of near-optimal policies.
\\\\
After the raw neural data has been pre-processed, we propose that some relationship exists between features of the neural data and the learning task statistics outlined above. We let $
\hat{y} = \phi(x_i, h_i)
$ denote the relationship between a vector of human neural data at some instance and its optimality label.

\section{Methodology}
\vspace{-5pt}
\textbf{Participants}
We recruited 25 participants for a partially within-participants study. Participants were between 19-27 years old and were recruited through physical and virtual fliers in a university setting and online.

\textbf{Equipment} $\vert$ Neural data was collected using an ISS OxiplexTS fNIRS device. This device uses pulsating infrared lasers to calculate the change in hemodynamic response under the human skull. The data was collected and pre-processed to account for motion artifacts and noise. Three OpenAI Gymnasium reinforcement learning domains were adjusted for seamless interfacing with participants: Robot Fetch and Place, Lunar Lander and Flappy Bird.

\textbf{Domains} $\vert$ An agent $\mathcal{A}$ may operate in one of three OpenAI Gymnasium's Reinforcement Learning environments. The Robot environment offered the following continuous actions: \texttt{x-axis}, \texttt{y-axis}, \texttt{z-axis}, \texttt{gripper}. The Lunar Lander environment offered the following discrete actions: \texttt{up}, \texttt{left}, \texttt{right}, \texttt{down}. The Flappy Bird environment offered the following discrete actions: \texttt{up}, \texttt{down}. 

\textbf{Experiment Procedure} $\vert$ Participants were given an informed consent form, configured to the fNIRS device, and instructed on the task. Then, they completed the task, and responded to two post-experiment questionnaires which included a standard NASA-TLX. 

Participants completed three or four out of six tasks dependent on time as to not exceed 60 minutes of procedure. These tasks lasted anywhere from 2-5 minutes each. Tasks were randomly selected, and could be any combination of a domain (3) and some task condition (2), resulting in six unique tasks. Task conditions were either \textit{passive} or \textit{active}. \textit{Passive} tasks instructed participants to observe and reflect on the performance of an autonomous agent interacting within its environment. \textit{Active} tasks instructed participants to physically guide an agent towards its goal state through the use of a keyboard or joystick. Participants are seated 24 inches from a computer screen, and a 20-second pause calibrates the fNIRS device before the task begins. 

\textbf{Passive Condition} $\vert$ The \textit{passive} condition, or "watch" demonstration, instructed participants to observe an agent achieve some goal in a given domain. The autonomous agent initially selects actions from a near-optimal policy and can be described as successful. With probability $p$, the agent transitions to a sub-optimal action-selection state, becoming unsuccessful during that episode. In an environment with a discrete action space, sub-optimal actions are selected by altering the probability distribution to select a sub-optimal action. In an environment with a continuous action space, some magnitude of noise, or the agent's goal state was altered. In our experiments, $p$ was roughly equivalent to $0.3$.

The agent continues to navigate the environment until the end of the task. The neural data, learning tasks statistics, and timestamps are stored in a hash-map and saved as a de-identified 'watch demonstration'.

\textbf{Active Condition} $\vert$ The \textit{active} condition, or "play" demonstration, instructs a participant to attempt the task themselves, guiding an agent towards some goal in a given domain. Participants used an Xbox controller to control the robot, and a basic computer keyboard to guide Flappy Bird and Lunar Lander agents. At the beginning of the task, the participant had to learn how to navigate the agent, but the goal was made clear. As they play, a one-hot vector was saved, representing the participant's selected action.

During the task, the participant's neural data, learning task statistics, and timestamps are stored in a hash-map for each episode and saved as a whole, de-identified 'play demonstration'.

\section{Machine Learning Approaches}
\vspace{-5pt}
\textbf{Pre-Processing and Feature Extraction} $\vert$ The raw fNIRS data was calibrated using the 20-second baseline at the beginning of each trial. The raw data was pre-processed using a 4th order Butterworth filter. Pre-processed data includes two features of the neural signals including phase and intensity values of the left and right pre-frontal cortex.

We derive a feature vector $\mathbf{F} \in \mathbb{R}^{L \times T}$ from the pre-processed neural signals, where $T$ is the set of timestamps and $L$ is the number of features. These features include mean, standard deviation, slope, intercept, skewness and kurtosis, resulting in a vector of length $8 \cdot x$, where $x$ is the number of features used. Each feature vector corresponds to an optimality label.

\textbf{Sliding Window Approach for Time Classification} $\vert$ We use a sliding window approach that extracts a fixed duration of $x$ seconds, where each window is assigned a single label. This label is defined based on the unique, final endpoint timestamp associated with the window. The endpoint of each window signifies the desired moment for evaluating the fNIRS data. We use a window large enough to encompass average latency associated with the hemodynamic response: approximately 5-7 seconds.

We denote our time series as:
$
x^{i,p}_{1:T} = [x^{i,p}_{1}, x^{i,p}_{2}, \dots, x^{i,p}_{T}]
$
where each subject $i$ has a series of $T$ vectors, $x$, each terminating at an endpoint $p$. Each vector $x^{i,p}_t \in \mathbb{R}^D$ represents data at time $t$ within a total of $T$ timesteps. The endpoint $p$ is associated with a label $y^{i,p} \in \mathcal{Y}$.

\textbf{Machine Learning Algorithms} $\vert$ We train a Support-Vector Machine (SVM), K-Nearest Neighbors (KNN), Decision Tree, Random Forest and Multilayer Perceptron (MLP) to determine which method produces the most valuable model. Each model is trained and tested on a set of participants from a specific condition. Models are then tested on data from different conditions to determine transferability without fine-tuning.

\textbf{Classifiers} $\vert$ To identify action optimality in both classifiers and regressors, both discrete, binary and continuous classification labels are saved. These labels are determined by the \emph{degree of failure} and the \emph{point of failure}. The degree of failure is an error value that determines how sub-optimal a selected action was. The point of failure is the moment within the episode where an agent begins to fail, usually a few seconds before the sparse reward indicates a lost episode.

\textbf{Multi-Policy Action Agreement} $\vert$ To calculate these values, we've designed a multi-policy action classification system. In any reinforcement learning problem, there may be more than one optimal path to the same goal. To avoid labeling agent behavior incorrectly, we compare the agent's chosen action with 10 near-optimal policies. An error value is calculated between the agent's action and each near-optimal policies action before taking an average. Error is calculated using Kullback-Leibler (KL) Divergence:
$
E_k(t) = D_{\text{KL}} \left( a_t \; \| \; \pi_k(s_t) \right) = \sum_{i=1}^n a_{t,i} \log \left( \frac{a_{t,i}}{\pi_k(s_{t,i})} \right)
$.
\\
\textbf{Binary Labels} $\vert$ Binary optimality labels are denoted as $\mathbf{B}_k \in \mathbb{R}^{T}$. At some time step $t$, $\mathbf{B}_k$ may be some binary number from the set
$ B_t = \{0,1\} $ where $0$ represents optimal behavior and $1$ represents sub-optimal behavior. In an episode where an agent fails, the point of failure is calculated and each feature vector afterwards is labeled sub-optimal.

\textbf{Discrete Labels} $\vert$ Discrete optimality labels are denoted as $\mathbf{V}_k \in \mathbb{R}^{T}$. At some time step $t$, $\mathbf{V}_k$ may be some discrete number from the set
$ V_t = \{0,1,2\} $ where $0$ represents optimal behavior, $1$ represents sub-optimal behavior and $2$ represents opposite, or worst-case behavior. In an episode where an agent fails, the point of failure and degree of failure is calculated and each feature vector afterwards is labeled accordingly.

\textbf{Continuous Labels} $\vert$ Optimal play errors can be denoted as $\mathbf{E}_k \in \mathbb{R}^{m}$
At some time step $t$, $\mathbf{E}_k$ may be some continuous value where lower values indicate near-optimal play, and higher values indicate sub-optimal to worst-case play.  The error between the agent's action \( a_t \) and the \( k \)-th near-optimal policy's action \( \pi_k(a_t) \) can be defined using the error function mentioned.

\section{Results}
\vspace{-5pt}
Our results revealed that classical machine learning algorithms can predict agent performance from fNIRS signals with greater success than random chance. Random Forest, MLP and KNN models generally performed best when solving classification problems. Random Forest and KNN generally outperformed other models when solving regression problems. Surprisingly, discrete models performed similarly to binary models (MLP Discrete-Class Across Conditions: $72.2\%$, MLP Binary-Class Across Conditions: $76.7 \%$). However, further analysis shows that discrete models struggled slightly with differentiating sup-optimal performance from optimal and worst-case (Discrete Optimal - AVG: $76\%$, Discrete Sub-Optimal - AVG: $67\%$, Discrete Worst-Case - AVG: $72\%$)

\begin{figure}[t]
\centering
\subfigure{\includegraphics[width=0.7\textwidth]{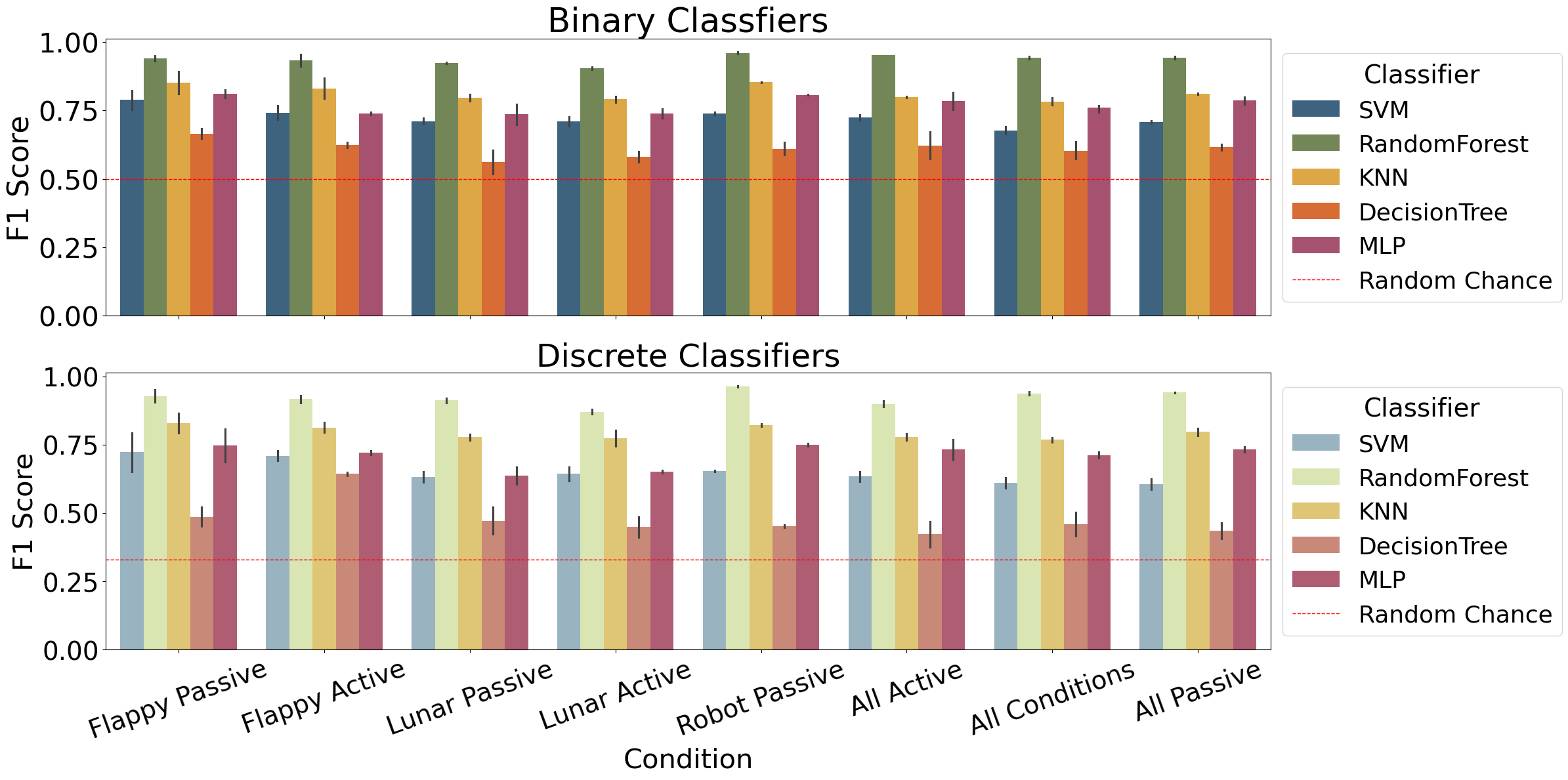}}
\caption{Classification Model Performance: Target class F1 score across models for each task condition.} \label{fig:1}
\end{figure}
\vspace{-5pt}

Feature type also made a significant difference in model performance. Simplifying the feature vector to mean, skewness, and standard deviation allowed some models, like MLP, to dramatically increase performance (Non-optimized Features F1 Score: $~72\%$, Optimized Feature F1 Score: $86.3\%$). We believe this is partially due to lowering the dimensionality of the feature vector, however mean and skewness were crucial to high performance compared to input vectors lacking these features.

\begin{figure}[h]
\includegraphics[width=16cm]{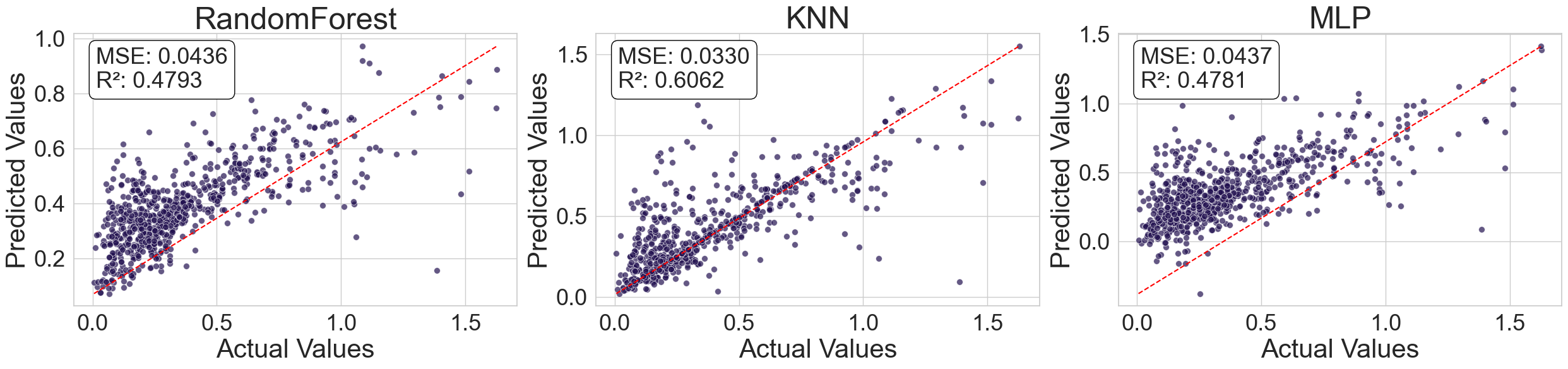}
\centering
\caption{\textmd{Regression Performance: Actual values and predicted values for each regressor (Lunar Lander Passive).}}
\end{figure}
\vspace{-5pt}
\section{Conclusion}
\vspace{-5pt}
Our study provides evidence that agent performance can be extracted from human neural data during some interaction. Most models struggled with cross-participant and cross-domain transferability, a common limitation in Brain-Computer Interface work. Other limitations include potential artifacts from fNIRS latency and noise, participant inattention, and dataset imbalances. Therefore, future work should focus on three major avenues. First, we hope to improve generalization by employing deep learning and data balancing techniques. Further, we plan to explore ways to handle signal latency, noise and participant inattention. Finally, we plan to apply this signal to a real-time RL framework and test its ability to implicitly fine-tune agent behavior online. 

We believe that reinforcement learning from neural feedback has undiscovered potential in fields like human-robot interaction, assistive AI, and various autonomous systems. Implicitly aligning agent behavior with human goals, expectations and preferences may change the way people interact with and perceive artificial and robotic agents in the future.
\printbibliography
\end{document}